\documentclass[lettersize,journal]{IEEEtran}
\usepackage{amsmath,amsfonts}
\usepackage{algorithmic}
\usepackage{algorithm}
\usepackage{array}
\usepackage[caption=false,font=normalsize,labelfont=sf,textfont=sf]{subfig}
\usepackage{textcomp}
\usepackage{stfloats}
\usepackage{url}
\usepackage{verbatim}
\usepackage{graphicx}
\usepackage{cite}

\usepackage{bbding}
\usepackage{amsmath}
\usepackage{amssymb}
\usepackage{booktabs} 
\usepackage{orcidlink}
\usepackage{soul}
\usepackage{makecell}
\usepackage{colortbl}
\usepackage{xcolor}
\usepackage{multirow}

\begin{document}

\title{WE-GS: An In-the-wild Efficient 3D Gaussian Representation for Unconstrained Photo Collections}


\author{Yuze Wang\orcidlink{0009-0000-7676-3408}, Junyi Wang\orcidlink{0000-0002-3191-1662}, Yue Qi\orcidlink{0000-0001-9304-1933}}

\markboth{Journal of \LaTeX\ Class Files,~Vol.~14, No.~8, August~2021}%
{Shell \MakeLowercase{\textit{et al.}}: A Sample Article Using IEEEtran.cls for IEEE Journals}


\maketitle

\begin{abstract}

Novel View Synthesis (NVS) from unconstrained photo collections is challenging in computer graphics. Recently, 3D Gaussian Splatting (3DGS) has shown promise for photorealistic and real-time NVS of static scenes. Building on 3DGS, we propose an efficient point-based differentiable rendering framework for scene reconstruction from photo collections. Our key innovation is a residual-based spherical harmonic coefficients transfer module that adapts 3DGS to varying lighting conditions and photometric post-processing. This lightweight module can be pre-computed and ensures efficient gradient propagation from rendered images to 3D Gaussian attributes. Additionally, we observe that the appearance encoder and the transient mask predictor, the two most critical parts of NVS from unconstrained photo collections, can be mutually beneficial. We introduce a plug-and-play lightweight spatial attention module to simultaneously predict transient occluders and latent appearance representation for each image. After training and preprocessing, our method aligns with the standard 3DGS format and rendering pipeline, facilitating seamlessly integration into various 3DGS applications. Extensive experiments on diverse datasets show our approach outperforms existing approaches on the rendering quality of novel view and appearance synthesis with high converge and rendering speed.
Project Page:\href{ https://yuzewang1998.github.io/we-gs.github.io/}{ https://yuzewang1998.github.io/we-gs.github.io/}

\end{abstract}


\section{Introduction}

\begin{figure*}[htbp]
 \includegraphics[width=1.0\linewidth]{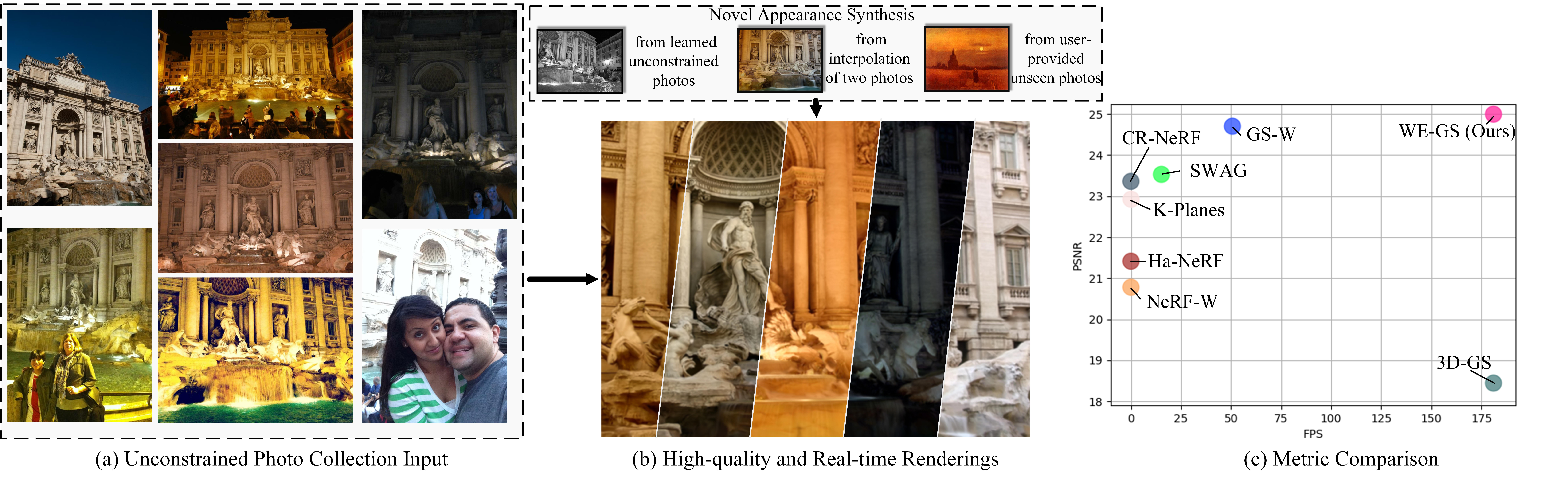}
 \centering
  \caption{(a) Given an unconstrained photo collection with appearance variation and transient occluders, (b) our method demonstrates the capability for real-time, high-quality novel view and appearance synthesis. (c) In comparison with previous and concurrent methods, our WE-GS framework can achieve the best rendering quality and real-time rendering performance while maintaining fast training speed. Photos by Flickr users /\href{https://creativecommons.org/licenses/by/2.0/}{CC BY}. }
\label{fig:teaser}
\end{figure*}
\IEEEPARstart{R}{econstructing} real-world scenes from photo collections with achieving photo-realistic Novel View Synthesis (NVS) is vital but challenging task in the fields of computer graphics and computer vision. NVS plays a significant role in numerous applications, including virtual reality, 3D content generation, and autonomous driving. In particular, Neural Radiance Fields (NeRF) \cite{nerf_vanilla} has inspired many follow-up works which have shown impressive progress in various aspects such as better visual effects \cite{mip_nerf360,tri-miprf}, faster rendering speed \cite{mobile_nerf,merf}, faster converging speed \cite{ingp}, more controllable editing capability \cite{nerf-editing,rip-nerf}, etc. More recently, 3D Gaussian Splatting (3DGS) \cite{3dgs} proposed to represent scene as 3D Gaussians and achieved high-fidelity and real-time rendering results.

Although NeRF and 3DGS demonstrate great NVS capability with well-captured photo collections of static scenes, capturing dense and high-quality photo collections for outdoor scenes, especially in renowned locations like the Brandenburg Gate in Berlin, remains challenging. 
To this end, some methods develop to broaden the application scope of NeRF for reconstructing scene with large-scale Internet photo collections of tourist landmarks, also known as in-the-wild scene reconstruction. In these unconstrained photo collections, input images may have been taken years apart and include the transient occluders, such as pedestrians and vehicles, moving through them. Efficiently and accurately modeling the appearance variation and predicting the transient occlusions for each unconstrained image during the 3D scene reconstruction process are the core challenges. 
NeRF-W \cite{nerf-w} was the first work to learn NeRF from unconstrained photo collections. NeRF-W optimized an appearance embedding and a transient radiance field for each unconstrained image with the Generative Latent Optimization (GLO) techniques. Followed by NeRF-W, a series of works \cite{ha-nerf,cr-nerf,refinedfields,k-planes} improve the prediction of transient occluders by leveraging 2D visibility maps. They also integrate more advanced neural networks to predict latent appearance representations. However, the aforementioned NeRF-based methods still face the intrinsic computational burden associated with volumetric rendering. Most recently, SWAG \cite{swag} proposed to expand the capabilities of 3DGS with a learned embedding space that modulates the color of 3D Gaussians with a Multi-Layer Perceptron (MLP). 
Another intuitive method, GS-W \cite{gs-w}, introduced two extra attributes for each 3D Gaussian: intrinsic and dynamic appearance features, to enable the varying appearance modeling.
However, in SWAG and GS-W, the dense distribution of 3D Gaussians throughout the 3D space leads to a notable increase in memory storage, as each 3D Gaussian requires additional attributes. Furthermore, SWAG and GS-W are unable to maintain the impressive real-time rendering performance of vanilla 3DGS.

Although there are some initial works on in-the-wild scene reconstruction, the above approaches still suffered from the following two critical shortcomings: \textbf{1) Lack of time-space efficiency in the reconstruction and rendering processes}: In order to predict the variable illumination and transient occluders for each unconstrained image, additional learnable parameters and training strategy need to be introduced, leading to the convergence of the training process slower. Moreover, since more advanced 3D representation baselines, e.g., iNGP \cite{ingp}, TensoRF \cite{tensorf}, and 3DGS \cite{3dgs}, cannot be directly integrated into the in-the-wild scene reconstruction task, the existing in-the-wild reconstruction methods still cannot account for real-time rendering, fast training, and compact data storage, thus lacking in time-space efficiency; \textbf{2) Lack of accurate transient mask prediction and latent appearance representation for each unconstrained image}: Due to the absence of constraints, accurately estimating the transient mask and latent appearance representation for each image is a challenging task, directly undermining the quality of novel view or novel appearance rendering. How to accurately predict the transient mask and appearance representation remains an open question.

As shown in Fig. \ref{fig:teaser}, in order to address the above challenges, we propose WE-GS, an efficient framework for reconstructing scenes from unconstrained photo collections. Specifically, to enhance the time-space efficiency in the reconstruction and rendering process, we expand the capabilities of 3DGS and introduce a residual-based Spherical Harmonic (SH) coefficients transfer module to adapt 3DGS to variable lighting conditions. This module ensures the efficient propagation of the gradients from rendered images to the attributes of 3D Gaussians, maximally preserving the efficient differentiable property, fast training, and rendering speed of vanilla 3DGS. From another perspective, unlike existing methods that predict transient mask and latent appearance representation through two separate neural networks, we observe that the appearance encoder and transient mask predictor can be mutually beneficial. We design a lightweight spatial attention module to simultaneously predict the transient mask and the latent appearance embedding for each unconstrained image, enhancing the prediction accuracy of the transient mask and representative of the latent appearance embedding. Various experiments on the PhotoTourism dataset \cite{pt} and NeRF-OSR dataset \cite{nerf-osr} are conducted with the proposed WE-GS, and shown a new state-of-the-art for training speed, rendering frames per second (FPS), and novel view or novel appearance synthesis quality for in-the-wild scene reconstruction. 
Specifically, for the evaluation on the PhotoTourism dataset, WE-GS maintains rendering speed while achieving over a 2 $\times$ reduction in storage and an average 6.6 dB increase in PSNR, indicating improved rendering quality, compared with vanilla 3DGS. Overall, the contributions of our method are summarized as:

\begin{itemize}
    \item We introduce WE-GS, an efficient point-based differentiable rendering framework for reconstructing scenes from unconstrained photo collections, setting a new state-of-the-art NVS quality.
    \item We propose a residual-based SH transfer module that adapt 3DGS to variable lighting and photometric post-processing. This module is lightweight and precomputable, which preserves a running speed on par with the vanilla 3DGS with minimum overhead.
    \item We observe that the appearance encoder and the transient mask predictor, the two most critical components of in-the-wild scene reconstruction, can be mutually beneficial. We design a plug-and-play lightweight spatial attention module capable of predicting both more accurate transient mask and more representative latent appearance representation for each unconstrained image simultaneously.
    \item We demonstrate that the proposed WE-GS also achieves state-of-the-art performance across various novel appearance synthesis applications. Besides, after precomputing, WE-GS maintains the standard format and rendering pipeline of the vanilla 3DGS, allowing for seamless integration into a broader range of 3DGS applications.
\end{itemize}

\section{Related Work}
\begin{figure*}[htbp]
\centering
\includegraphics[width=2.0\columnwidth]{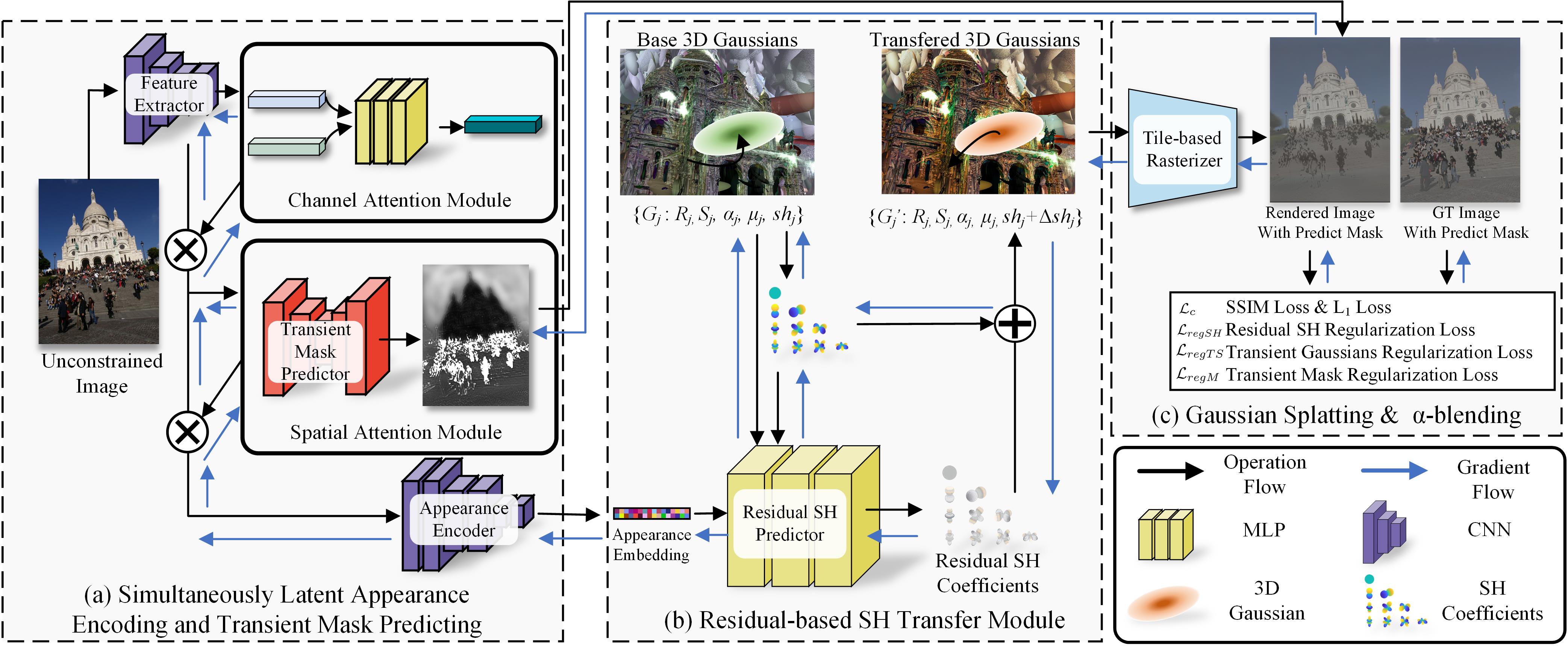}
\caption{\textbf{An overview of the proposed WE-GS framework.} (a) Given an unconstrained image, we propose a lightweight spatial attention module to simultaneously predict the transient mask and the latent appearance representation. (b) Then a residual-based SH coefficient transfer module is used to modeling the appearance variation. (c) Finally, we adopt the tile-based rasterizer to render the image and supervise it with several proposed regularization loss and the predicted transient mask to tackle the transient occluders in the unconstrained image.}
\label{method_main}
\end{figure*}
\subsection{Scene Representations and Radiance Fields}
Mesh \cite{mesh_1}, point cloud \cite{point_1}, volume \cite{volume_1}, and implicit function \cite{if_1} are traditional 3D scene representation methods. They have been extensively studied and applied in the fields of computer graphics and computer vision. In recent years, Neural Radiance Fields (NeRF) introduced radiance fields to address the Novel View Synthesis (NVS) task and achieved photo-realistic rendering results. Subsequently, a growing number of NeRF extensions have emerged, such as aiming to enhance visual effects \cite{mip_nerf360,tri-miprf}, improve rendering speed \cite{mobile_nerf,merf}, accelerate convergence \cite{ingp}, and provide more controllable editing capabilities \cite{nerf-editing,rip-nerf}, etc. More recently, 3DGS \cite{3dgs} has garnered attention from researchers due to its real-time rendering speed and photo-realistic rendering quality. Many researchers are exploring its potential by enhancing its performance, such as developing more compactness 3DGS representation \cite{gs_compact_1}, improving visual quality of 3DGS  \cite{gs_vis_mip-splatting,gs_vis_scaffold-gs}, or augmenting the capability of 3DGS with introducing additional semantic attributes \cite{gs_attr_1,gs_attr_2}. Moreover, 3DGS also widely extended to various applications, including Simultaneous Localization And Mapping (SLAM) \cite{gs_slam_gs-slam1}, AI-Generated Content (AIGC) \cite{gs_aigc_dreamgs}, or scene understanding \cite{goi}.

While 3DGS performs well on static scenes with well-captured photo collections, it struggles to model ubiquitous real-world phenomena in unconstrained photo collections, such as variable illumination and transient occluders.

\subsection{Neural Rendering from Unconstrained Photo Collections}
Reconstructing and performing NVS of a scene from unconstrained photo collections has long been a challenge in computer graphics and computer vision \cite{snavely2006photo}. NeRF brings a new dimension to address this task. NeRF-W \cite{nerf-w} was the first work to utilize NeRF for reconstructing scenes from unconstrained photo collections. It employed the Generative Latent Optimization (GLO) techniques to optimize an appearance embedding and a transient radiance field for each input unconstrained image. To enable novel appearance synthesis, Ha-NeRF \cite{ha-nerf} introduced a Convolutional Neural Network (CNN) based appearance encoder and utilized a view-consistent appearance loss to ensure consistent photometric appearance transfer across different unconstrained images. CR-NeRF \cite{cr-nerf} introduced a cross-ray paradigm to leverage global information across multiple rays, thereby enhancing rendering quality. K-Planes \cite{k-planes} proposed a volumetric NeRF representation using planar factorization along with an MLP decoder. This method aims to separate global appearance and enhance convergence speed. Very recently, SWAG \cite{swag} introduced an enhancement to 3DGS by incorporating a learned embedding space that modulates the color of 3D Gaussians using an MLP. Similarly, GS-W \cite{gs-w} introduced intrinsic and dynamic appearance features to each 3D Gaussian, enabling reconstruction from unconstrained photo collections. 

However, the above methods faced challenges in reconstructing scene details, achieving real-time rendering, and maintaining compact storage, which limited their further applications. 
\subsection{Spatial Attention Module}
It is widely recognized that a key characteristic of the human visual system is its selective processing of salient parts of a scene, rather than attempting to process the entire scene simultaneously. This selective focus allows for better capture of the visual structure. Drawing inspiration from this characteristic, spatial attention modules improve neural network performance by directing focus towards important features while suppressing unnecessary ones. CBAM \cite{sa0_cbam} introduced an attention module for convolutional neural networks, capable of attending to important information in channels and spatial locations separately. Following CBAM, its extensions have been widely applied to various vision tasks, including image classification \cite{sa_image_class}, image segmentation \cite{sa_image_seg}, and image detection \cite{sa_image_detection}. Motivated by the success of CBAM, we introduce a plug-and-play lightweight spatial attention module. This module simultaneously predicts the transient occluders mask and the latent appearance representation for each unconstrained image, thereby enhancing the performance of in-the-wild scene reconstruction. 

\section{Preliminaries}

In 3D Gaussian Splatting (3DGS) \cite{3dgs}, a scene is represented as millions of 3D Gaussians. Each 3D Gaussian $G_i$ is characterized by a set of learnable parameters, including the 3D center position $\mu_i \in \mathbb{R}^3$, the 3D covariance matrix $\Sigma_i \in \mathbb{R}^{3 \times 3}$, the opacity $\alpha_i \in \mathbb{R}^{+}$ and the Spherical Harmonic (SH) coefficients $sh_i \in \mathbb{R}^{L^2}$ for view-dependent appearance, where  $L$ is the degree of SH. The 3D Gaussians are defined in world space as follows:
\begin{equation}
G_{i}(x) = e^{-\frac{1}{2} (x-\mu_i)^T \Sigma_i^{-1}(x-\mu_i)}.
\end{equation}
The 3D covariance matrix $\Sigma_i$ is obtained using its rotation matrix $R_i \in \mathbb{R}^{3 \times 3}$ and corresponding diagonal scale matrix $S_i \in \mathbb{R}^{3 \times 3}$:
\begin{equation}
\Sigma_i = R_{i}S_{i}S_{i}^{T}R_{i}^{T}.
\end{equation}
Given the Jacobian of the affine projective transformation $J$ and the viewing transformation $W$, the 2D covariance matrix $\Sigma_i^{\prime}$ and 2D center $\mu_i^{\prime}$ are as follows:
\begin{equation}
\Sigma_i^{\prime} = JW\Sigma_i W^{T}J^{T},    
\end{equation}
\begin{equation}
\mu_i^{\prime} = JW\mu_i.
\end{equation}
These 3D Gaussians are projected to 2D splats and blended using the $\alpha$-blending process to obtain the rendered image.The color of pixel $u$, denoted as $C({u})$, can be obtained using $M$ ordered 2D splats with the following formula:
{\small
\begin{equation}
\label{equa:gaussian_render}
C ({u}) = \sum_{i \in M} T_i \alpha_i^{\prime} \mathcal{SH}(sh_i, d), \text{ where } T_i = \Pi_{j=1}^{i - 1}(1 - \alpha_{j}^{\prime}),
\end{equation}}
where $\mathcal{SH}$ is the SH function and $d$ is the view direction. $\alpha_i^{\prime}$ represents the final opacity of the 3D Gaussian, obtained by:
\begin{equation}
\label{equa:render2}
\alpha_i^{\prime} = \alpha_i e^{-\frac{1}{2} ({y} - \mu_i^{\prime})^T \Sigma_i^{\prime} ({y} - \mu_i^{\prime}) },
\end{equation}
where $y$ represents the corresponding coordinates in the projected space. By optimizing all parameters of the 3D Gaussians $\mathcal{G}=\{G_i: \mu_i, R_i, S_i, \alpha_i, sh_i  |  i=1,...,N\}$, a static scene can be reconstructed and photo-realistic novel views can be rendered in real-time.

While 3DGS performs well on photo collections of static subjects captured under controlled settings, it struggles to model ubiquitous real-world phenomena from unconstrained photo collections, such as variable illumination or transient occluders. 3DGS assumes that the radiance (i.e., SH coefficients $sh_i$) and the opacity $\alpha_i$ of the world are constant. However, even photos taken at the same time and position may exhibit different transient occluders, camera exposures, color corrections, and post-processing in unconstrained photo collections, leading to a significant performance degradation of 3DGS.
\section{Methodology}

Our goal is to efficiently and realistically reconstruct scenes from unconstrained photo collections, capturing appearance variations and transient occluders. In the following sections, we introduce WE-GS, an point-based differentiable rendering framework for in-the-wild scene reconstruction. We first demonstrate how 3DGS can be adapted to handle variable lighting and photometric post-processing using the proposed residual-based SH transfer module in Sec. \ref{method_sh}. Next, we introduce the lightweight spatial attention module, which simultaneously predicts the transient mask and the latent appearance representation for each unconstrained image in Sec. \ref{method_encoder}. Finally, we describe the optimization process in Sec. \ref{method_optimization}. An overview of our method is presented in Fig. \ref{method_main}.

\subsection{WE-GS Representation with Residual-based SH Transfer Module.}
\label{method_sh}
\begin{figure}[htbp]
\centering
\includegraphics[width=1.0\columnwidth]{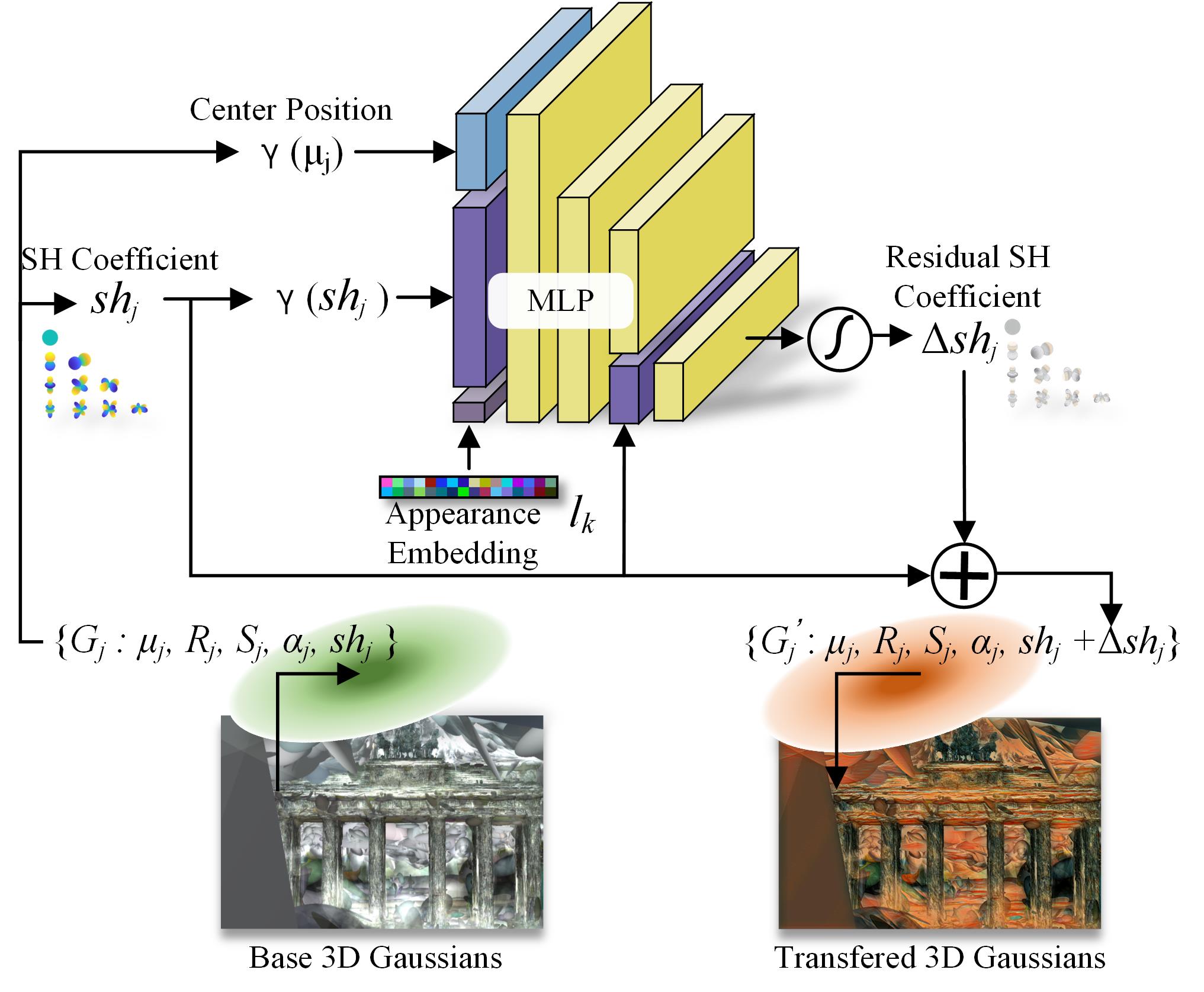}
\caption{\textbf{Residual-based SH transfer module.} Given the appearance embedding vector $l_k$, each 3D Gaussian $G_j$ learns image-sepcific residual SH coefficients $\Delta sh_{jk}$ with a lightweight MLP. By the summing of image-agnostic base SH coefficients $sh_{j}$ and image-specific residual SH coefficients $\Delta sh_{jk}$, appearance variable from unconstrained photo collections can be modeling.}
\label{figure_sh}
\end{figure}

\begin{figure*}[htbp]
\centering
\includegraphics[width=2.0\columnwidth]{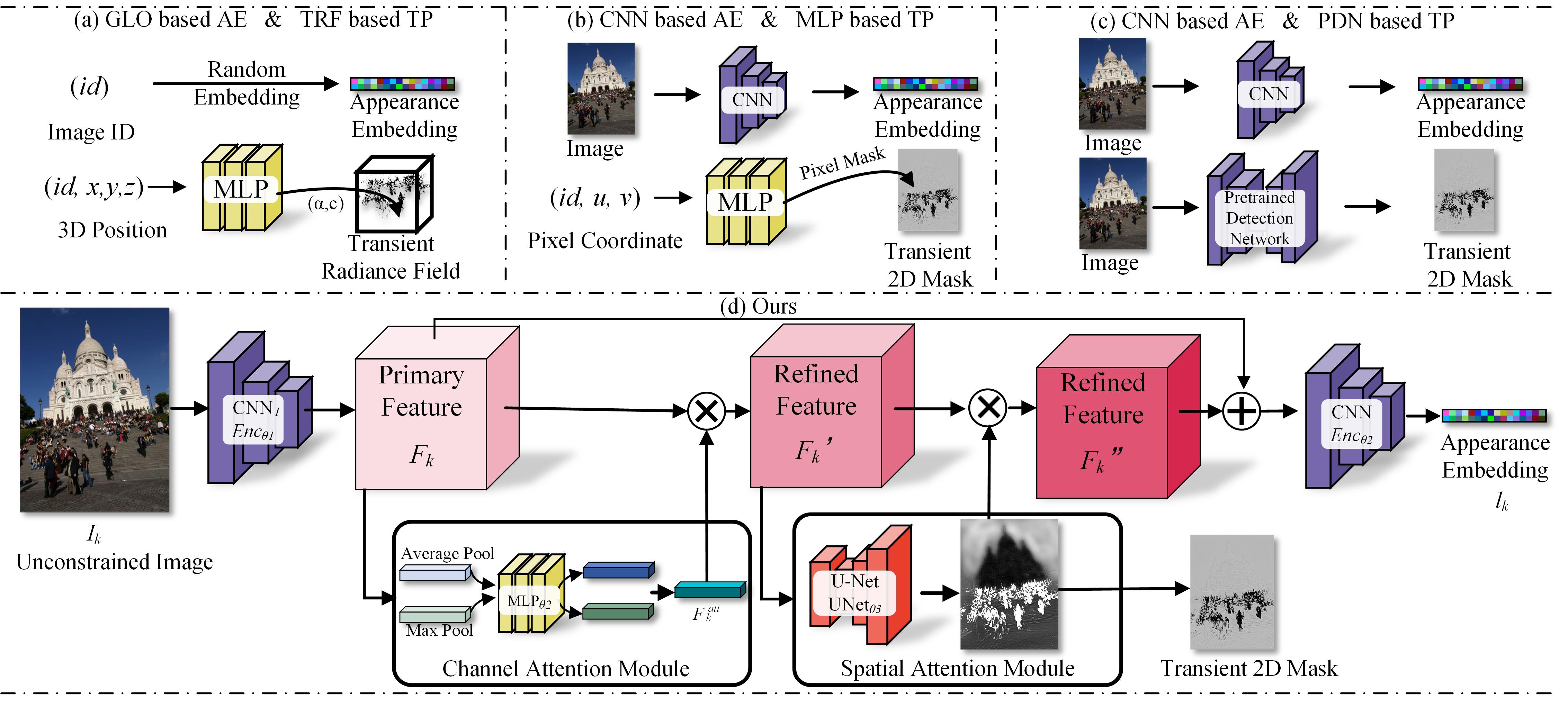}
\caption{\textbf{The appearance encoder and transient mask predictor.} (a), (b), and (c): Three types of existing methods are demonstrated to separately model the appearance embedding and predict the transient mask. (d): We find that the appearance encoder and the transient mask predictor can be mutually beneficial. We propose a plug-and-play lightweight spatial attention module to simultaneously predict the transient mask and the latent appearance embedding for each unconstrained image. }
\label{fig_method42}
\end{figure*}
To adapt 3DGS to handle variable lighting and photometric post-processing, we inject per-image appearance information into the 3D Gaussian representation. Given an unconstrained image $\mathcal{I}_k$, we learn an appearance embedding vector $l_{k}$ as follows:
\begin{equation}
 l_{k} = Enc_{\theta}(\mathcal{I}_k).
\end{equation}

Here, $Enc_{\theta}$ represents a neural network parameterized by $\theta$, which will be discussed in detail in Sec. \ref{method_encoder}. To learn a per-image specific radiance $sh_{jk}^{\prime}$ for each 3D Gaussian $\{G_j: \mu_j, R_j, S_j, \alpha_j, sh_j \}$, an intuitive solution proposed in GS-W \cite{gs-w} is to introduce additional attributes for 3D Gaussians and design a NeRF-like space-coordinate-based MLP to query the changes of SH at a specific 3D position:
\begin{equation}
    sh_{jk}^{\prime} = \mathcal{F_\eta}(\mu_j,f_j,l_k).
\end{equation}
In GS-W, $f_j$ denotes the fusion of two additional attributes introduced for the 3D Gaussian. However, this design has drawbacks. On the one hand, the inference and optimization stage is time-consuming, similar to standard NeRF, as it requires dense querying for every sampled 3D space coordinate and processing through multiple linear layers. On the other hand, such a design is space-consuming since it introduces additional attributes for millions of 3D Gaussians. To maintain the space-time efficiency of 3DGS, we propose learning an implicit mapping from the base SH coefficients to per-image specific SH coefficients, thereby modeling appearance variation. As demonstrated in Fig. \ref{figure_sh}, we employ an MLP to predict the residual SH coefficients $\Delta sh_{jk}$ for each 3D Gaussian from the appearance embedding vector $l_k$, SH coefficients $sh_{j}$, and center position $\mu_j$, i.e.:
\begin{equation}
\label{delta_sh_cauc}
    \Delta sh_{jk} = \mathcal{G}_\phi(\gamma(sh_j), \gamma(\mu_j), l_k),
\end{equation}
\begin{equation}
    sh_{jk}^{\prime} = sh_j + \Delta sh_{jk},
\end{equation}
where $\gamma(.)$ is the Positional Encoding (PE) function introduced in NeRF \cite{nerf_vanilla}.
This design is motivated by the assumption that for different unconstrained images $I_i$ and $I_j$, with their respective radiance fields $RF_i$ and $RF_j$, there exists a relationship between any two points $x$ and $y$ in 3D space:
\begin{equation}
    RF_i(x) - RF_j(x) \approx RF_i(y) - RF_j(y),
\end{equation}
where $RF_i(x)$ is the SH coefficients of 3D point $x$ in radiance field $RF_i$. Furthermore, as we know, $\Delta sh_j$ is strongly correlated with $sh_j$. Therefore, the neural network $\mathcal{G}_\phi$ in Eq. \ref{delta_sh_cauc}  can be kept shallow and lightweight. In our experiments, $\mathcal{G}_\phi$ is implemented as a four-layer MLP.

Another advantage of this design is that it effectively prevents the gradients of the attributes of each 3D Gaussian from vanishing or exploding, thereby ensuring the optimization efficiency of the parameters. For each 3D Gaussian, the gradient of the SH coefficients can be expressed as:

{\footnotesize
\begin{align}
    \frac{\partial C}{\partial \mathcal{SH}(sh_i,d)} & = T_{i}\alpha_{i}^{\prime}\frac{\partial \mathcal{SH}(sh_{i}+\mathcal{G}_\phi(\mathcal{SH}(sh_j,d), \mu_j, l_k), d)}{\partial \mathcal{SH}(sh_i,d)}    \nonumber \\
    & = \underbrace{ T_{i} \alpha_{i}^{\prime}}_{Vanilla\: 3DGS} + \underbrace{T_{i} \alpha_{i}^{\prime}\frac{\partial \mathcal{G_\phi}(\mathcal{SH}(sh_j,d),\mu_j,l_k)}{\partial \mathcal{SH}(sh_i,d)}.}_{Extra\: introduced\: gradient}
\end{align}}
Experimentally, we observed that the latter term is typically less impactful than the former. Compared to gradients obtained by the cumulative multiplication case, the proposed residual-based SH transfer module maximally ensures the optimization efficiency of vanilla 3DGS.

Besides, we found that although $sh_j+\Delta sh_{jk}$ can successfully model appearance variations, the base $sh_j$ tends to be offset, thus losing its physical meaning. Therefore, to address this issue, we propose a regularization loss term for $\Delta sh_{jk}$ of each 3D Gaussian $G_j$ and unconstrained image $\mathcal{I}_k$: 
\begin{equation}
\mathcal{L}_{regSH} = \sum_{j \in N, k \in M} \Vert \Delta sh_{jk} \Vert . 
\end{equation}
Here, $N$ represents the number of 3D Gaussians, and $M$ represents the size of the unconstrained image collection. In our experiments, we found that incorporating such a regularization term improves the scene reconstruction result.

Importantly, WE-GS can pre-transfer the SH coefficients of all 3D Gaussians for a given unconstrained image. As a result, for interactive rendering tasks, WE-GS maintains the running speed on par with vanilla 3DGS. Furthermore, after precomputing, WE-GS maintains the standard format and rendering pipeline of the vanilla 3DGS, allowing for seamless integration into a broader range of 3DGS applications.

\subsection{Simultaneously Transient Mask Predicting and Latent Appearance Encoding}
\label{method_encoder}

Handling the transient occluders with the Transient mask Predictor (TP) and encoding the appearance with the Appearance Encoder (AE) for each unconstrained image are two key components for in-the-wild scene reconstruction. The accurate prediction of the transient mask and latent appearance representation directly determines the quality of the NVS. 
At the top of Fig. \ref{fig_method42}, the three existing AE and TP methods are illustrated. As shown in top-left of Fig. \ref{fig_method42}, NeRF-W \cite{nerf-w} and SF-NeRF \cite{sf-nerf} utilized a Generative Latent Optimization (GLO) based AE for appearance encoding and predict the transient mask using an additional Transient Radiance Field (TRF). Since GLO  based AE does not consider the color features of the unconstrained image itself, it can only interpolate between already learned appearances.
As demonstrate in the top-middle of the Fig. \ref{fig_method42}, Ha-NeRF \cite{ha-nerf}, GS-W \cite{gs-w} and SWAG \cite{swag} proposed a CNN based AE to extract more representative features from unconstrained images, enabling encoding of untrained images for novel appearance synthesis. For TP, they all learned the transient mask in 2D pixel space, where an MLP, given image identifier $id$ and pixel coordinates $(u,v)$, learns the corresponding 2D transient mask. However, this approach tends to 'mechanically memorize' rather than 'generalize learning', which negatively impacts the accuracy of transient mask predictions. Alternatively, CR-NeRF \cite{cr-nerf} fine-tuned a Pre-trained Detection Network (PDN) \cite{cg-net} to predict transient 2D mask, as shown in the top-right of Fig. \ref{fig_method42}. However, such AE increases the number of parameters for training and storage and lack of generalization, as it requires specifying categories (e.g., people, cars) for detection and masking in advance.

Unlike existing methods, we observed a mutually beneficial relationship between AE and TP. We propose an unsupervised learning approach to predict transient mask and appearance representation simultaneously. As illustrated in the bottom of Fig. \ref{fig_method42}, we propose guiding the appearance encoder to focus more on buildings and backgrounds with the assistance of the transient mask predictor.
Specifically, given an unconstrained image $\mathcal{I}_k \in \mathbb{R}^{3 \times W \times H}$, a CNN encoder is used to extract the primary feature $F_k \in \mathbb{R}^{C \times W^{\prime} \times H^{\prime}}$, i.e. :

\begin{equation}
     F_k = Enc^1_{\theta1}(\mathcal{I}_k).
\end{equation}
We begin by constructing a channel attention module to exploit the inter-channel relationship of features. Each channel in $F_k$ serves as a feature detector for the input unconstrained image. Then, we aggregate spatial information of the features using both average-pooling and max-pooling operations. These pooled feature are then passed through a small shared-weight MLP $MLP_{\theta2}$ to generate the 1D channel attention map $F_{k}^{c}  \in \mathbb{R}^{C \times 1 \times H^{\prime}} $:

{\footnotesize
\begin{equation}
    F_{k}^{c} = Sig(MLP_{\theta2}(AvgPool(F_k)) + MLP_{\theta2}(MaxPool(F_k))).
\end{equation}}

We apply the Sigmoid activation function $Sig(.)$ to the channel attention map. The resulting values are then used to perform element-wise multiplication with the original feature map $F_k$, yielding the refined feature $F^{\prime}_k$.
\begin{equation}
   F^{\prime}_k = F_{k}^{c}  \otimes F_k,
\end{equation}
where $\otimes$ denotes element-wise multiplication broadcasted along the spatial dimension. Channel attention module assigns different weights to different channels so the neural network can learn which layer is more important. Subsequently, we introduce an U-Net style spatial attention module to learn the static content of the unconstrained image:
\begin{equation}
   M_k =  {UNet}_{\theta3}(F^{\prime}_k),
\end{equation}
where $M_k \in \mathbb{R}^{1 \times W^{\prime} \times H^{\prime}}$ represents the static content of the scene, as well as the content that one hopes to focus the appearance encoder's attention on. The complement of $M_k$ is the predicted transient 2D mask, denoted as $\mathcal{R} (1-M_k) \in \mathbb{R}^{1 \times W \times H}$, where $\mathcal{R}$ represents the resize operation to align the shape with the input image $\mathcal{I}_k$. The output static content mask, also referred to as the spatial attention feature, focuses on identifying 'where' informative parts of the feature are located. Then, we fuse the spatial attention feature with the refined feature map $F^{\prime}_k$ using the following equation to obtain the refined feature $F^{\prime\prime}_k \in \mathbb{R}^{C \times W^{\prime} \times H^{\prime}}$:
\begin{equation}
   F^{\prime\prime}_k = M_k \otimes F^{\prime}_k
\end{equation}
Finally, by incorporating a residual connection and employing a CNN encoder $Enc^2_{\theta4}$, we achieve the appearance embedding $l_k$ of the unconstrained image $\mathcal{I}_k$:
\begin{equation}
  l_k = Enc^2_{\theta4}(F^{\prime\prime}_k + F_k).
\end{equation}

\subsection{Optimization}
\label{method_optimization}
The proposed WE-GS framework is optimized in an end-to-end manner. We simultaneously optimize the parameters of all the 3D Gaussians  $\mathcal{G}=\{G_i: \mu_i, R_i, S_i, \alpha_i, sh_i  |  i=1,...,N\}$, the residual-based SH transfer module $\mathcal{G}_{\phi}$, the latent appearance encoding and transient mask predicting module $\{Enc_{\theta1}^1, MLP_{\theta2}, UNet_{\theta3}, Enc_{\theta4}^2\}$. Following to vanilla 3DGS \cite{3dgs}, we apply the L1 loss $\mathcal{L}_1$ and Structural Similarity Index (SSIM) \cite{ssim} loss $\mathcal{L}_{SSIM}$ to measure the error between the rendered image $\mathcal{I}_k^{r}$ and the ground truth (GT) image $\mathcal{I}_{k}^{gt}$. However, unlike 3DGS, we mask out the transient occluders by the predicted static content mask $M_k$:

{\footnotesize
\begin{equation}
    \mathcal{L}_{c} = \lambda_1 \mathcal{L}_1( M_k \otimes \mathcal{I}_k^{r},M_k \otimes \mathcal{I}_k^{gt} )+
                        \lambda_{ssim} \mathcal{L}_{SSIM}( M_k \otimes \mathcal{I}_k^{r},M_k \otimes \mathcal{I}_k^{gt} ),
\end{equation}}

where $\otimes$ is the pixel-wise multiplication, and $\lambda_1$, $\lambda_{ssim}$ denote the weights assigned to the L1 loss and SSIM loss, respectively. To prevent the transient mask predictor from excessively masking, we incorporate a regularization term following Ha-NeRF \cite{ha-nerf}:
\begin{equation}
    \mathcal{L}_{regM} = \| 1- M_k \|^{2}
\end{equation}
In addition, while well-capture photo collections can produce high-quality point clouds for the initializing 3D Gaussians with the help of the Structure from Motion (SfM) techniques, such as COLMAP \cite{colmap}. However, the point clouds generated from unconstrained photo collections tend to be noisy, with many of the noisy points representing transient occluders. Utilizing such noisy point clouds as the initialization for 3D Gaussians can increase storage overhead and negatively affects the efficiency of the optimization. We observe that as the accuracy of transient mask prediction improves, the opacity gradients of these noisy 3D Gaussians become minimal. Therefore, we introduce a new constraint term aimed at penalizing the opacity values of those 3D Gaussians falling within regions of 2D transient masks after projection. This constraint term serves to mitigate the impact of noisy initialization on optimization performance, i.e.:
\begin{equation}
    \mathcal{L}_{regTS} = \sum_{i \in N} \alpha_i \mathbb{I}(M(\Pi(\mu_i))< \epsilon) ,    
\end{equation}
where $\mathbb{I}$ is the indicator function, $\Pi$ is the projection function, and $\epsilon$ is a hyper-parameter. When $\alpha_i$ is sufficiently small, we follow the vanilla 3DGS to prune that Gaussian. The final loss function can be formalized as following:
\begin{equation}
    \mathcal{L} = \mathcal{L}_{c} +\lambda_1 \mathcal{L}_{regM} +\lambda_2 \mathcal{L}_{regSH} +\lambda_3 \mathcal{L}_{regTS},
\end{equation}
where $\lambda_1$, $\lambda_2$, and $\lambda_3$ are the hyperparameter used to balance the contributions of different loss functions. 

\section{Experiments}
\begin{table*}[htbp]
    
    \caption{Quantitative experimental results on three scenes in PhotoTourism dataset \cite{pt} compared to previous works. WE-GS demonstrates state-of-the-art performance on the task of reconstruction from unconstrained photo collections.}
    \label{tab:quantitative_compare_pt}
    \centering
    \resizebox{1.7\columnwidth}{!}{%
    \begin{tabular}{lcccccccccc}
        
        \cmidrule[\heavyrulewidth]{2-10}
         & \multicolumn{3}{c}{Bradenburg Gate}  & \multicolumn{3}{c}{Sacre Coeur} & \multicolumn{3}{c}{Trevi Fountain}   \\
        \cmidrule(rl){2-4}
        \cmidrule(rl){5-7}
        \cmidrule(rl){8-10}
        
         & PSNR $\uparrow$  & SSIM$\uparrow$  & LPIPS $\downarrow$ & PSNR $\uparrow$  & SSIM$\uparrow$  & LPIPS $\downarrow$ & PSNR $\uparrow$  & SSIM$\uparrow$  & LPIPS $\downarrow$ \\
        \midrule

        NeRF-W \cite{nerf-w}         
        & 24.17   & 0.890 & 0.167
           
            & 19.20 & 0.807 &0.191
           
            & 18.97  & 0.698 & 0.265\\ 
        
        HA-NeRF \cite{ha-nerf} 
        & 24.04   & 0.877 & 0.139
           
            & 20.02 & 0.801 &0.171
           
            & 20.18  & 0.690 & 0.222\\   
        
        CR-NeRF \cite{cr-nerf} 
        &  \cellcolor{yellow!25}26.53   & 0.900 & \cellcolor{orange!25}0.106
           
            & 22.07 & 0.823 &\cellcolor{yellow!25} 0.152
           
            & 21.48  & 0.711 & \cellcolor{yellow!25}0.206\\ 

        RefinedFields \cite{refinedfields}
         & 26.64   & 0.886 & -
           
            & \cellcolor{yellow!25}22.26 & 0.817 & -
           
            &\cellcolor{orange!25} 23.42  & 0.737 & -\\ 
         
        K-Planes \cite{k-planes} 
            &25.49  & 0.924 & -
        & 20.61 &  0.852 & -
        & 22.67 &\cellcolor{red!25} 0.856& -\\

        SWAG \cite{swag} 
            &26.33  & \cellcolor{yellow!25}0.929 & 0.139
        & 21.16 & \cellcolor{yellow!25} 0.860& 0.185
        & \cellcolor{yellow!25}23.10 &  \cellcolor{yellow!25} 0.815& 0.208\\

        GS-W \cite{gs-w} 
            &\cellcolor{red!25}27.96  & \cellcolor{orange!25}0.931 & \cellcolor{red!25}0.086
        & \cellcolor{orange!25}23.24 & \cellcolor{orange!25} 0.863& \cellcolor{red!25}0.130
        & 22.91 &   0.801& \cellcolor{orange!25}0.156\\

        3DGS \cite{3dgs} 
            &20.72  & 0.889 & 0.152
        & 17.57 &  0.839& 0.190
        & 17.04 &   0.690 & 0.265\\

        WE-GS (Ours) 
            &\cellcolor{orange!25}27.74  &\cellcolor{red!25}0.933 & \cellcolor{yellow!25}0.128
        & \cellcolor{red!25}23.62 &  \cellcolor{red!25}0.890&\cellcolor{orange!25} 0.148
        & \cellcolor{red!25}23.63 &  \cellcolor{orange!25} 0.823&\cellcolor{red!25} 0.153\\

        \bottomrule
    \end{tabular}
    }
\end{table*}
\begin{table*}[htbp]
        \caption{Quantitative experimental results on four scenes of NeRF-OSR dataset \cite{nerf-osr} compared to previous works. WE-GS outperformed the other works on PSNR, SSIM, LPIPS across all datasets.}
    \label{tab:quantitative_compare_nerf-osr}
    \resizebox{2.0\columnwidth}{!}{%
    \renewcommand{\arraystretch}{0.5}
    \setlength{\tabcolsep}{0.0035\linewidth}
    \centering
    \begin{tabular}{lccccccccccccc}
        \cmidrule[\heavyrulewidth]{2-13}
         & \multicolumn{3}{c}{europa}  & \multicolumn{3}{c}{lwp} & \multicolumn{3}{c}{st} 
         & \multicolumn{3}{c}{stjohann} \\
        \cmidrule(rl){2-4}
        \cmidrule(rl){5-7}
        \cmidrule(rl){8-10}
        \cmidrule(rl){11-13}
        
             & PSNR~$\uparrow$  & SSIM~$\uparrow$  & LPIPS~$\downarrow$ & PSNR~$\uparrow$  & SSIM~$\uparrow$  & LPIPS~$\downarrow$ & PSNR~$\uparrow$  & SSIM~$\uparrow$  & LPIPS~$\downarrow$ & PSNR~$\uparrow$  & SSIM$~\uparrow$  & LPIPS~$\downarrow$ \\
                
        \midrule
        NeRF~\cite{nerf_vanilla}

        &17.49 &0.551 &0.503&
        11.51& 0.468 &0.574 &
        17.20& 0.514 &0.502 &
        14.89 &0.432 &0.639 &\\

        NeRF-W~\cite{nerf-w}

        &20.00 &0.699& 0.340&
        19.61& 0.616 &0.445&
        {20.31} &0.607& 0.438& 
        {21.23} &0.667 &0.426 \\

        Ha-NeRF~\cite{ha-nerf}
        &17.79 &0.632& 0.421&
        {20.03}&{0.685}& {0.365} &
        17.30& 0.538& 0.483& 
        17.19 &0.686& 0.331\\

        NeRF-MS~\cite{nerf-ms} ~
        &\cellcolor{yellow!25}{21.03} &{0.721} &{0.294}&
        \cellcolor{yellow!25}{21.90} &\cellcolor{yellow!25}{0.719}& \cellcolor{yellow!25}{0.336}&
        \cellcolor{yellow!25}{20.68} &\cellcolor{yellow!25}{0.630}& \cellcolor{yellow!25}{0.402} &
        \cellcolor{yellow!25}{22.84} &\cellcolor{yellow!25}{0.793} &\cellcolor{yellow!25}{0.235}\\

        3DGS~\cite{3dgs} 
        &{20.18} &\cellcolor{yellow!25}{0.782}& \cellcolor{yellow!25}{0.252}&
        11.76 &0.609& 0.414&
        17.16 &{0.629}& {0.406} &
        16.77 &{0.741} &{0.268}&\\

        SWAG \cite{swag} 
           &\cellcolor{orange!25}{23.91} &\cellcolor{orange!25}{0.864} &\cellcolor{orange!25}{0.172}&
        \cellcolor{orange!25}{22.07} &\cellcolor{orange!25}{0.783}& \cellcolor{orange!25}{0.303}&
         \cellcolor{orange!25}{22.29} & \cellcolor{orange!25}{0.713}&  \cellcolor{orange!25}{0.364} &
        \cellcolor{orange!25}{23.74} &\cellcolor{orange!25}{0.845} &\cellcolor{orange!25}{0.242}\\
        \midrule
       WE-GS (Ours) 
           &\cellcolor{red!25}{24.73} &\cellcolor{red!25}{0.873} &\cellcolor{red!25}{0.157}&
        \cellcolor{red!25}{24.33} &\cellcolor{red!25}{0.821}& \cellcolor{red!25}{0.197}&
         \cellcolor{red!25}{22.45} & \cellcolor{red!25}{0.720}&  \cellcolor{red!25}{0.341} &
        \cellcolor{red!25}{24.12} &\cellcolor{red!25}{0.858} &\cellcolor{red!25}{0.202}\\
        \bottomrule        \\

    \end{tabular}
    }
    \end{table*}
\begin{table}[htb]
\caption{Quantitative comparison on the efficiency of our method, computed over three scenes in PhotoTourism dataset \cite{pt}. 
Results marked with dagger $\dagger$ have been directly adopted from the original paper.
"TT", "RT", and "ST" are "Training Time ", "Rendering Time", and "Storage", respectively.}
\label{tab:quantitative_compare_efficient}
\resizebox{1.0\columnwidth}{!}{
\begin{tabular}{lccccc}
\cmidrule[\heavyrulewidth]{2-6}
                              &TT (h) $\downarrow$ & RT (FPS) $\uparrow$ &ST (MB) $\downarrow$  & PSNR (dB) &\\ \hline
HA-NeRF \cite{ha-nerf}        & 35.1               & 0.07                &\cellcolor{red!25}17.4                       & 21.41     & \\
CR-NeRF \cite{cr-nerf}        & 31.0               & 0.04                &\cellcolor{orange!25}53.9                       & \cellcolor{yellow!25}23.36     & \\
K-Planes \cite{k-planes}      & \cellcolor{red!25}0.3                & 0.13                &425.3                      & 22.92&          \\
3DGS \cite{3dgs}     & \cellcolor{orange!25}0.4     & \cellcolor{red!25}181   &      140.8    &18.44  &              \\
SWAG \cite{swag}&  \cellcolor{yellow!25} 0.8$^\dagger$        & 15.29$^\dagger$ &     -             & 23.53$^\dagger$&          \\
GS-W \cite{gs-w}& 2.0$^\dagger$        & \cellcolor{yellow!25}50.70$^\dagger$ &     -             &\cellcolor{orange!25}24.70$^\dagger$&          \\

\hline
WE-GS (Ours)                       & 1.8   & \cellcolor{red!25}181   &    \cellcolor{yellow!25}  66.4    &\cellcolor{red!25}25.00 &      \\ \hline
\end{tabular}}

\label{table_nerf_editing_compare}
\end{table}

\begin{figure*}[htbp]
\centering
\includegraphics[width=2.0\columnwidth]{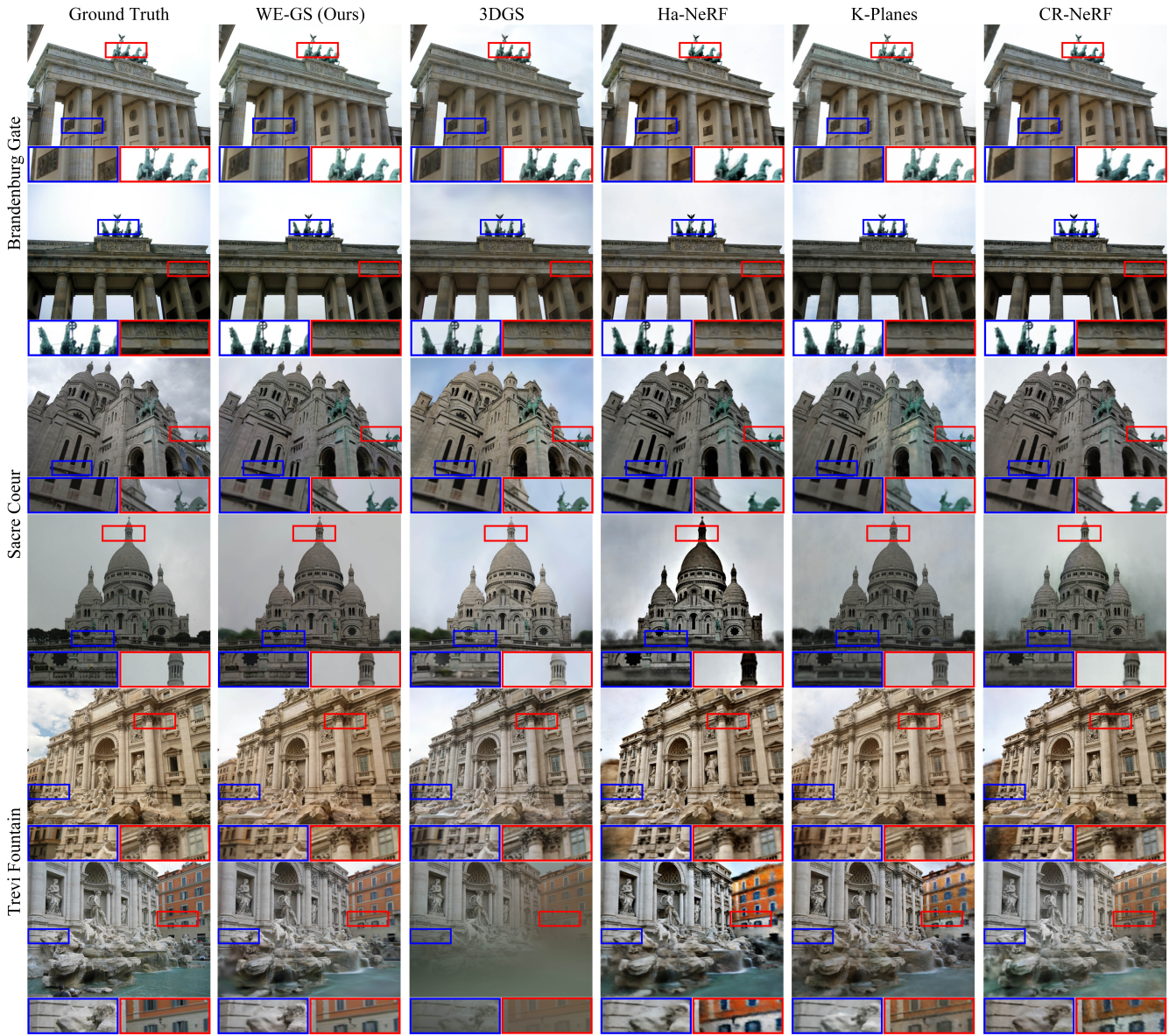}
\caption{Qualitative experimental results on three unconstrained real-world scenes from PhotoTourism dataset \cite{pt}. We compare our method with sate-of-the-art methods including 3D-GS \cite{3dgs}, Ha-NeRF \cite{k-planes}, K-Planes \cite{k-planes}, and CR-NeRF \cite{cr-nerf}.
Non-obvious differences in quality highlighted by insets.}
\label{exp_quati_pt}
\end{figure*}
\subsection{Implementation Details}
We implement our method in Python using the Pytorch framework \cite{pytorch} and train the proposed neural networks with Adam optimizer \cite{adam}. 
Both the appearance encoder $Enc_{\theta1}^1$ and $Enc_{\theta4}^2$ consist of 3 convolution layers, while $MLP_{\theta2}$ comprises 2 fully-connected layers followed by ReLU \cite{relu} activation function. The U-Net $UNet_{\theta3}$ comprises 2 convolution layers and 2 transpose layers. We set the dimension of appearance embedding $l_k$ to 16, and set $W^{\prime}$ and $H^{\prime}$ to be $1/4$ $W$ and $H$, respectively. Training was conducted on a single Nvidia RTX 4090 GPU for 150,000 steps.
Hyperparameter settings include $\lambda_{1}$ set to 0.8, $\lambda_{SSIM}$ set to 0.2, $\lambda_{1}$ set to $1 \times 10^{-5}$, $\lambda_{2}$ set to 0.1, and $\lambda_{3}$ set to $1 \times 10^{-5}$. Additionally, adaptive control over the 3D Gaussians was applied, following similar settings as in 3DGS \cite{3dgs}.
\subsection{Datasets and Metrics} 
Similar to previous in-the-wild reconstruction methods, we evaluated the NVS capabilities of WE-GS on three scenes in PhotoTourism dataset \cite{pt}: Brandenburg Gate, Sacre Coeur, and Trevi Fountain. Additionally, we also evaluate the proposed WE-GS on the outdoor scene relighting benchmark introduced in NeRF-OSR \cite{nerf-osr} to demonstrate the robustness of the proposed method.
We evaluate NeRF-OSR dataset across 4 scenes: \textit{europa}, \textit{lwp}, \textit{st}, and \textit{stjohann}.

We conducted comprehensive evaluations of WE-GS, including visual comparisons with rendered images and quantitative assessments based on PSNR, SSIM \cite{ssim}, and LPIPS \cite{lpips}. In addition to assessing performance, we also investigated the efficiency of our approach by comparing training time, rendering FPS, and storage footprint of the model parameters.

\subsection{Quantitative Comparison}

Quantitative results are shown in Tab. \ref{tab:quantitative_compare_pt} across three scenes in PhotoTourism dataset \cite{pt}. WE-GS improves an average $6.6$dB PSNR compared to the vanilla 3DGS \cite{3dgs}. 
We found that when reconstructing scenes from unconstrained photo collections, 3DGS performs worst among all methods. Notably, methods like NeRF-W and K-Planes cannot directly transfer appearance to untrained test images, requiring optimization of appearance embedding for each test image, potentially biasing the comparison. 
Despite with the biasing comparsion WE-GS consistently outperforms other methods. Tab. \ref{tab:quantitative_compare_nerf-osr} showcases the quantitative result on NeRF-OSR \cite{nerf-osr} dataset. We also reach state-of-the-art performance across multiple metrics. 
Specifically, WE-GS improves the average PSNR by $7.4$ dB compared to vanilla 3DGS and by $0.9$ dB compared to SWAG.

Tab. \ref{tab:quantitative_compare_efficient} demonstrates the qualitative comparisons involving training time, rendering time, storage size, and rendering quality. Training time and storage sizes are average across the above three scenes in PhotoTourism \cite{pt} datasets, while rendering time is average across all test sets in the above three scenes. As SWAG and GS-W have not made their code publicly available, we directly report their metrics from their original papers. WE-GS demonstrates high efficiency, requiring only 66.4 MB of space and enabling reconstruction in less than 1.8 hours. This corresponds to more than $17 \times$ speedup compared to NeRF-based methods.
 
\subsection{Qualitative comparison}

\begin{figure*}[htbp]
\centering
\includegraphics[width=1.9\columnwidth]{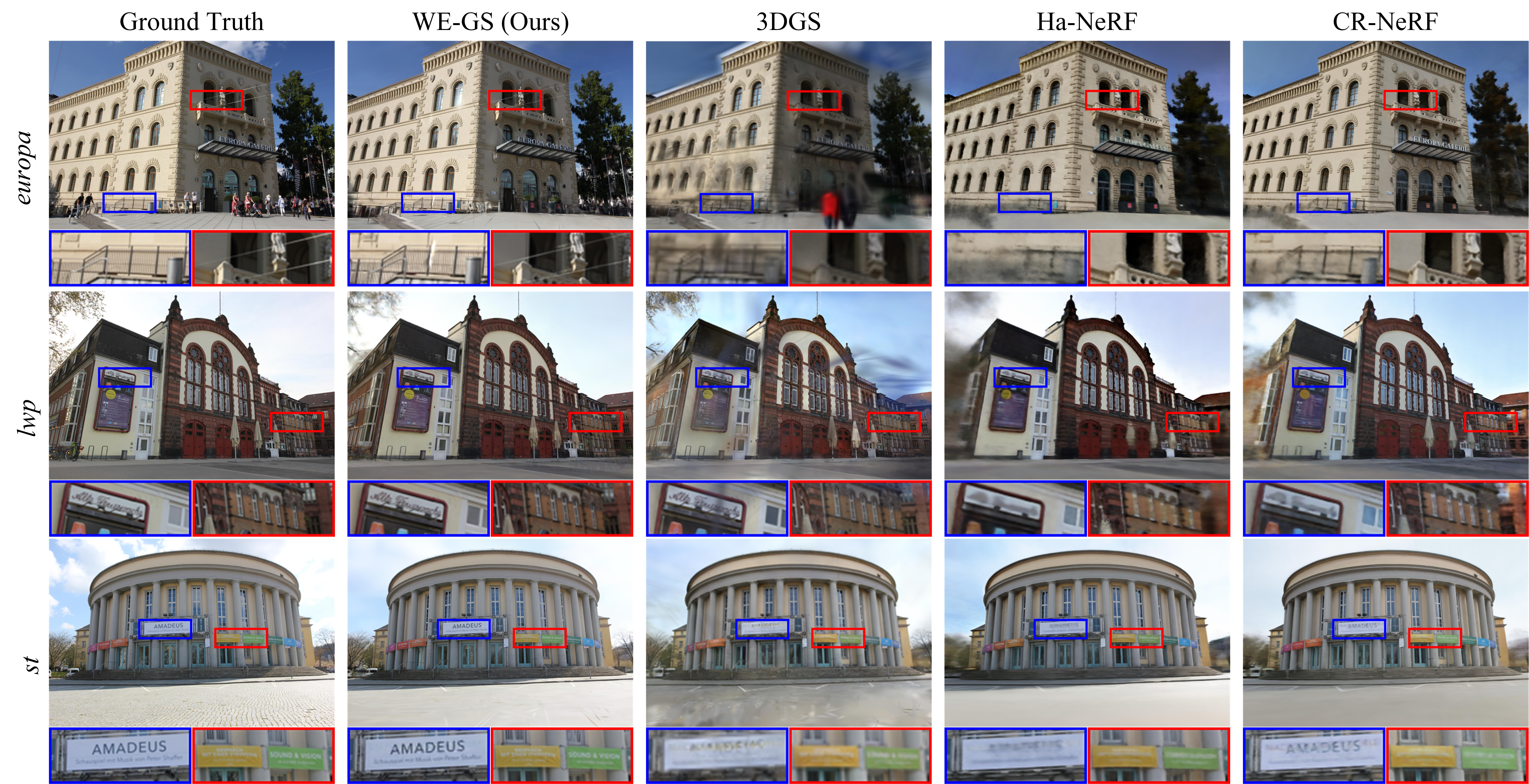}
\caption{Qualitative experimental results on 4 unconstrained real-world scenes from NeRF-OSR \cite{nerf-osr} dataset. We compare our method with sate-of-the-art methods including 3D-GS \cite{3dgs}, Ha-NeRF \cite{k-planes}, and CR-NeRF \cite{cr-nerf}.
Non-obvious differences in quality highlighted by insets.}
\label{exp_quati_nerf-osr}
\end{figure*}

\begin{figure}[htbp]
\centering
\includegraphics[width=1.0\columnwidth]{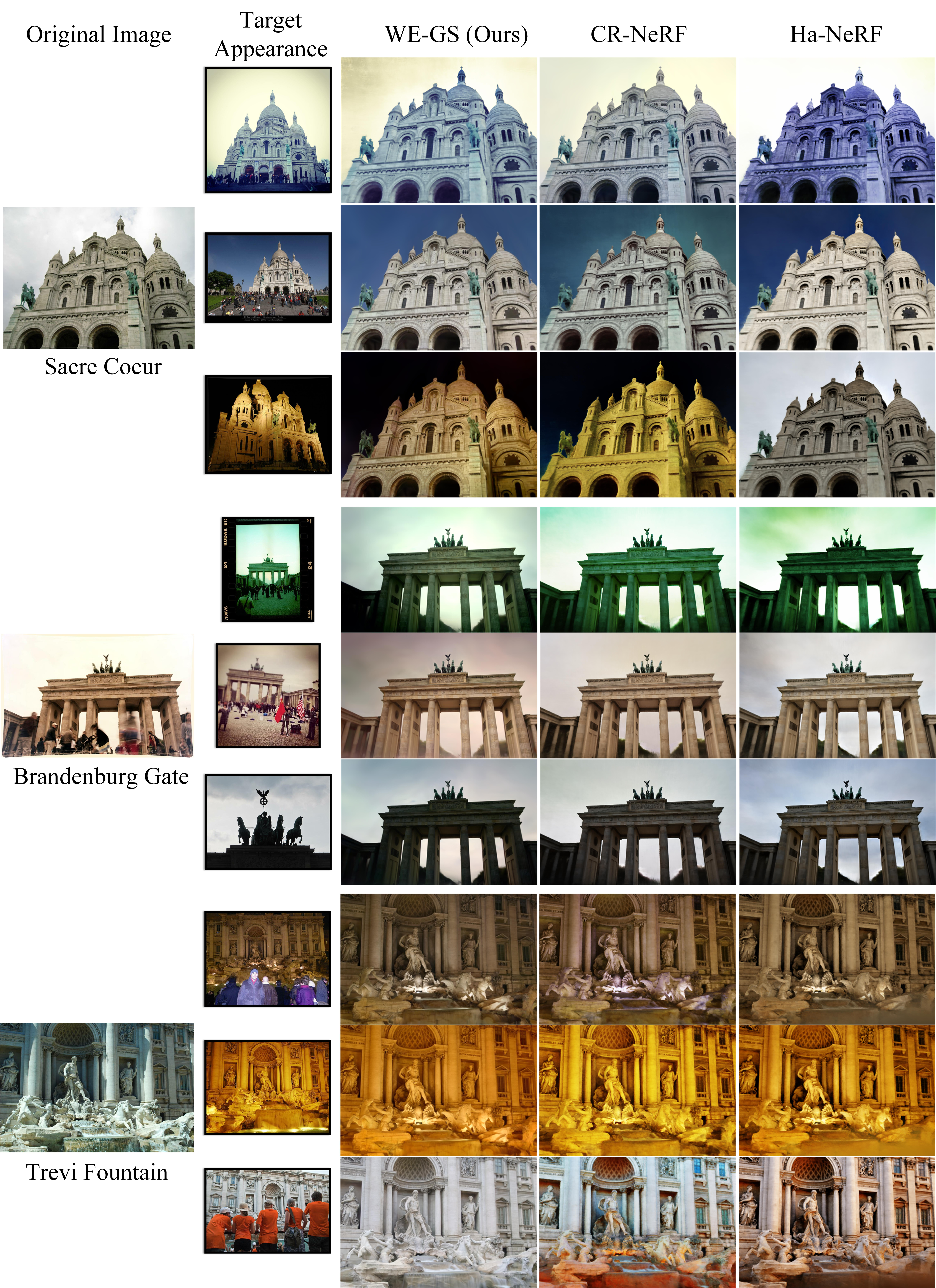}
\caption{Appearance modeling comparison results on PhotoTourism dataset \cite{pt}. WE-GS can predict and remove the transient occluders and recover the realistic appearance compared to Ha-NeRF \cite{ha-nerf} and CR-NeRF \cite{cr-nerf}.}
\label{exp_am_compare}
\end{figure}

\begin{figure*}[htbp]
\centering
\includegraphics[width=1.7\columnwidth]{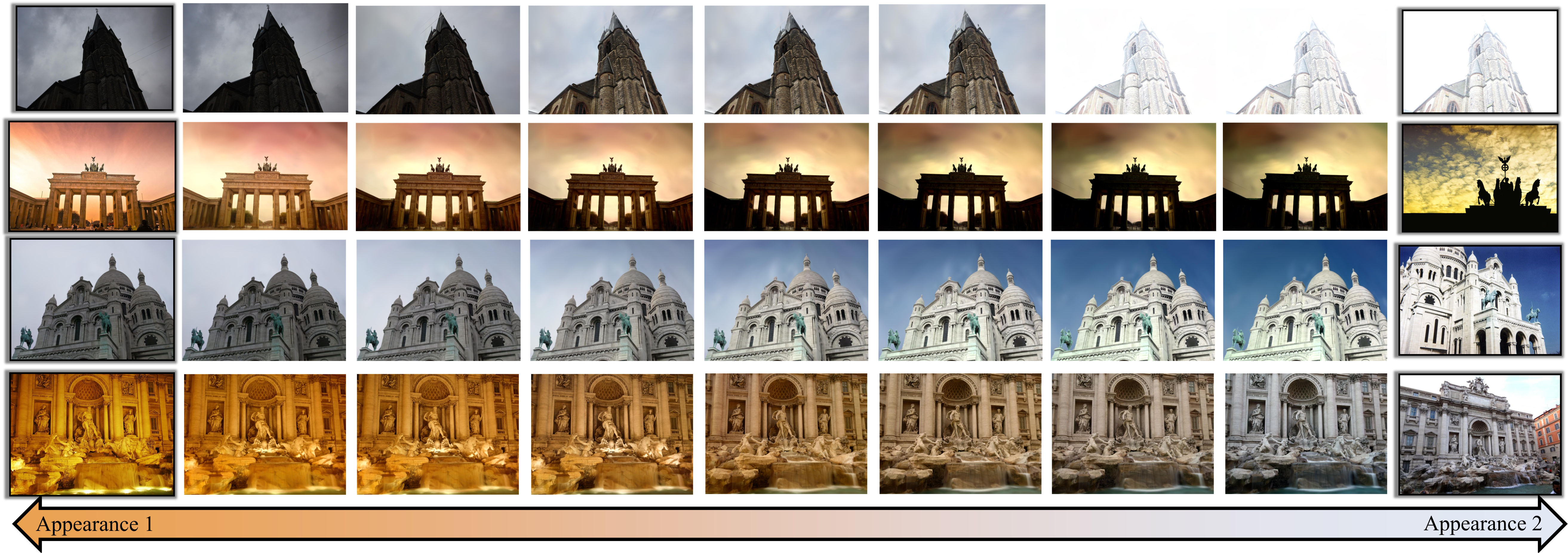}
\caption{Novel appearance synthesis results. Images are rendered from a fixed camera pose with the appearance embeddings interpolated between appearance 1 and appearance 2.}
\label{exp_am_interpolation}
\end{figure*}

\begin{figure}[htbp]
\centering
\includegraphics[width=1.0\columnwidth]{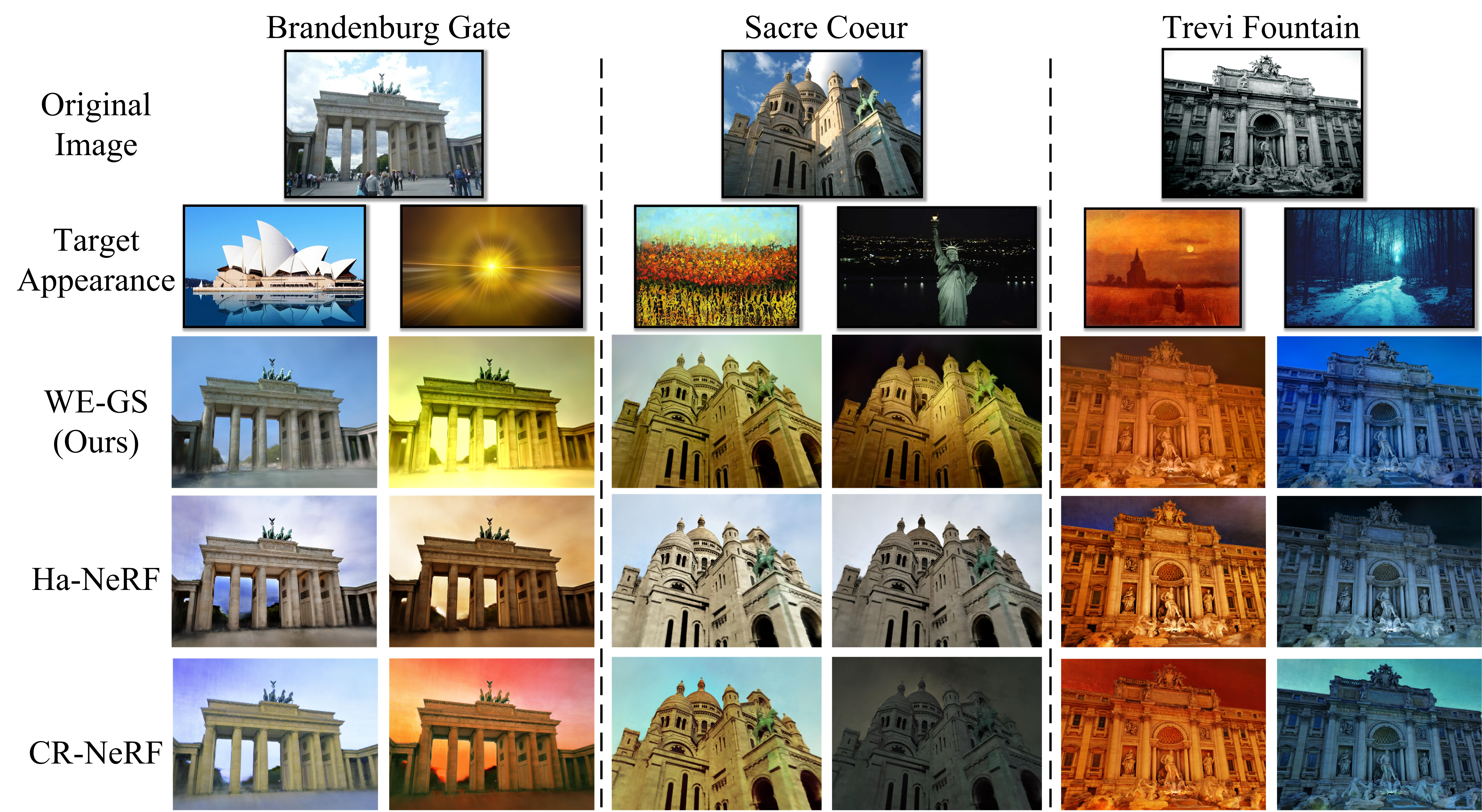}
\caption{Novel appearance synthesis with an arbitrary user-provided image. WE-GS can capture the style of the unseen image and transfer the scene to a more consistent appearance than Ha-NeRF \cite{ha-nerf} and CR-NeRF \cite{cr-nerf}. }
\label{exp_am_hal}
\end{figure}

We also conduct qualitative comparisons to showcase the advantages of our proposed method. The comparisons on PhotoTourism dataset are illustrated in Fig. \ref{exp_quati_pt}. 
Our method excels in capturing finer details of the scenes, such as the reins of the bronze wagon in Brandenburg and the sword of the knight sculpture in Sacre Coeur. In Fig. \ref{exp_quati_nerf-osr}, we present qualitative comparisons conducted on the NeRF-OSR dataset. 3DGS exhibits ghosting artifacts and global color shifts when reconstructing scenes from unconstrained photo collections. While CR-NeRF shows improved appearance variation effects compared to Ha-NeRF and K-Planes, it struggles to accurately reconstruct scene details. We strongly encourage the reader to inspect the supplemental video or project page for more comparison.
\subsection{Appearance Modeling Comparison}

Fig. \ref{exp_am_compare} showcases the appearance modeling capability of the proposed WE-GS compared to Ha-NeRF and CR-NeRF. WE-GS demonstrates superior accuracy in modeling appearance variations. Notably, in the last row of Fig. \ref{exp_am_compare}, we present a specific example. The target image features multiple people wearing orange tops. CR-NeRF and Ha-NeRF treat all pixels equally, encoding transient occluders as part of the scene's appearance, resulting in an incorrect orange color appearance in the rendered scene. However, thanks to the spatial attention module, the appearance encoder in WE-GS prioritizes non-transient regions, enabling precise disentanglement and modeling of the scene's appearance. We urge the reader to visually inspect the video results in the supplement materials or project page to observe more results with comparison.

\subsection{Appearance Interpolation}
We-GS has the capability to interpolate the appearance embedding between any two arbitrary images, enabling novel appearance rendering. In Fig. \ref{exp_am_interpolation}, we present three cases rendered from a fixed camera position, where we interpolate the appearance embedding encoded from the leftmost and the rightmost images. We demonstrate in the first line the results of interpolating images with different exposures. In rows 2-4, we exhibit the results of interpolation for various environment lighting variations. We also urge the reader to inspect the video supplement materials or project page for more results.

\subsection{Unseen Appearance Transfer}
WE-GS can also achieve appearance transfer given an arbitrary user-provided style image. By setting all the output values of the transient mask predictor to $0$, the appearance encoder does not give extra spatial attention to any pixel. As illustrated in Fig. \ref{exp_am_hal}, the user-provided style image exhibits a significant domain gap with the appearance of the learned scene. Despite this, we are able to capture a more accurate latent appearance from the style image compared to Ha-NeRF and CR-NeRF, showcasing the robustness of the proposed WE-GS.
\subsection{Ablation Studies}
\begin{table}[htbp]
\caption{Ablation studies of our method. Metrics are averaged over 3 scenes in PhotoTourism dataset \cite{pt}.}
\resizebox{1.0\columnwidth}{!}{
\begin{tabular}{lcccc}
\hline
                  & PSNR$\uparrow$           & SSIM$\uparrow$   & LPIPS$\downarrow$           & Size (MB)$\downarrow$      \\ \hline
(1) Dim of $l_i$ = 8       & 23.91        & 0.865          & 0.155  & 66.3\\
(2) Dim of $l_i$ = 32       & 24.87          & 0.876          & 0.146& 66.1                   \\
(3) Degree of $sh_j$ = 3       & 24.90          & 0.879          & 0.144& 45.9                   \\
(4) Degree of $sh_j$ = 5       & 24.95          & 0.881          & 0.144& 87.8                    \\\hline
(5) w/o residual SH $\Delta sh_{jk}$ & 23.81          & 0.871          & 0.149 & 66.4                    \\
(6) w/o CAM       & 24.96         & 0.878          & 0.145 & 66.2                   \\
(7) w/ MLP based TP     & 23.51          & 0.868          & 0.160& 63.6                   \\
(8) w/ CNN based TP       & 24.94          & 0.874          & 0.145  & 66.3                    \\\hline
(9) w/o $\mathcal{L}_{regSH}$ & 24.07          & 0.877          & 0.146   & 66.4                   \\
(10) w/o $\mathcal{L}_{regTS}$& 24.96         & 0.883          & 0.142   &  147.6                   \\\hline 
(11) Complete model     & 25.00 & 0.882 & 0.143 & 66.4  \\ \hline
\end{tabular}
}

\label{ablation_study}
\end{table}

\begin{figure}[htbp]
\centering
\includegraphics[width=1.0\columnwidth]{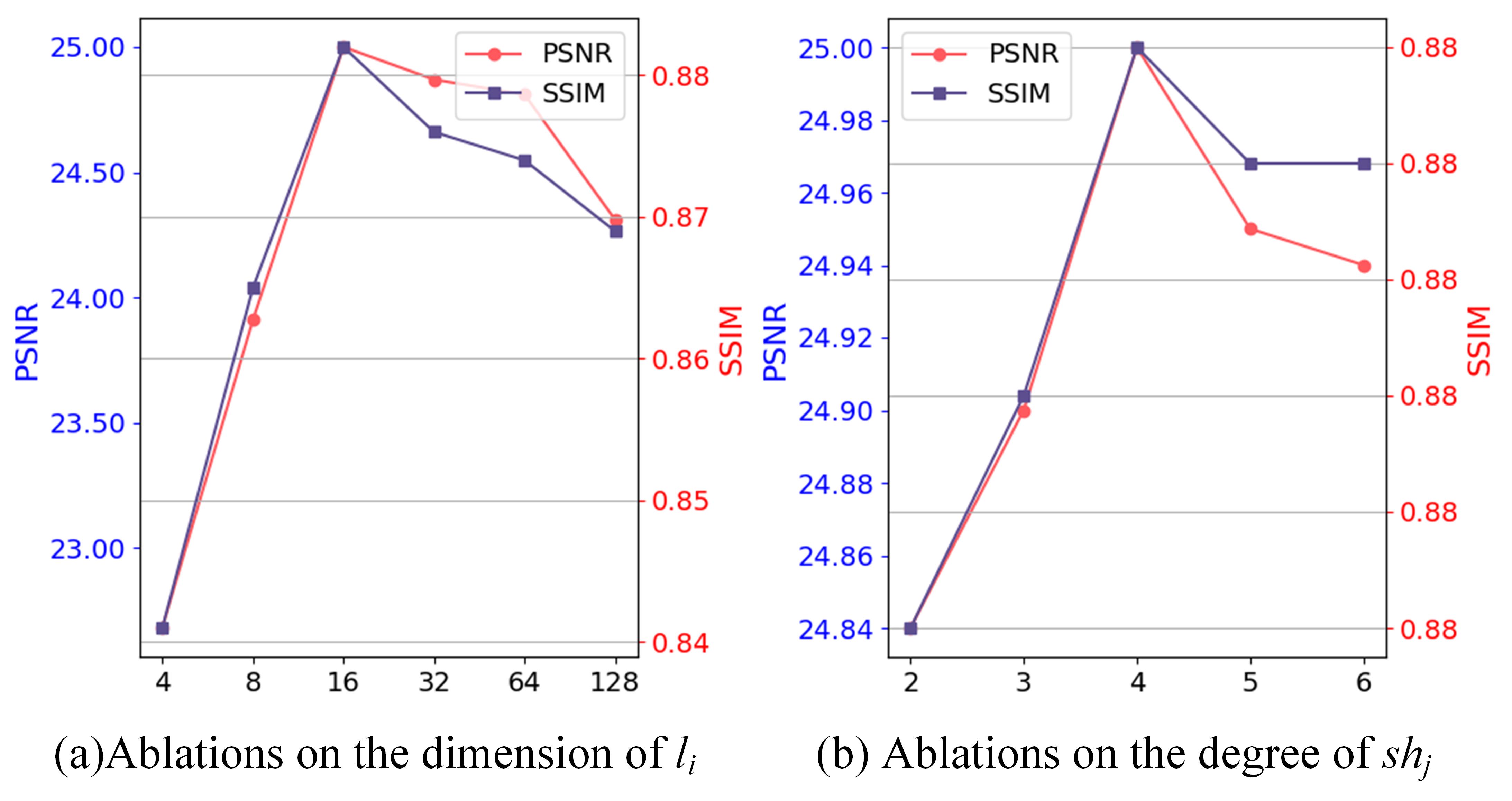}
\caption{Ablations on the dimention of the appearance embeddings $l_i$ (a) and the degree of the SH $sh_{j}$.}
\label{ablation_dim}
\end{figure}
\begin{figure}[htbp]
\centering
\includegraphics[width=0.88\columnwidth]{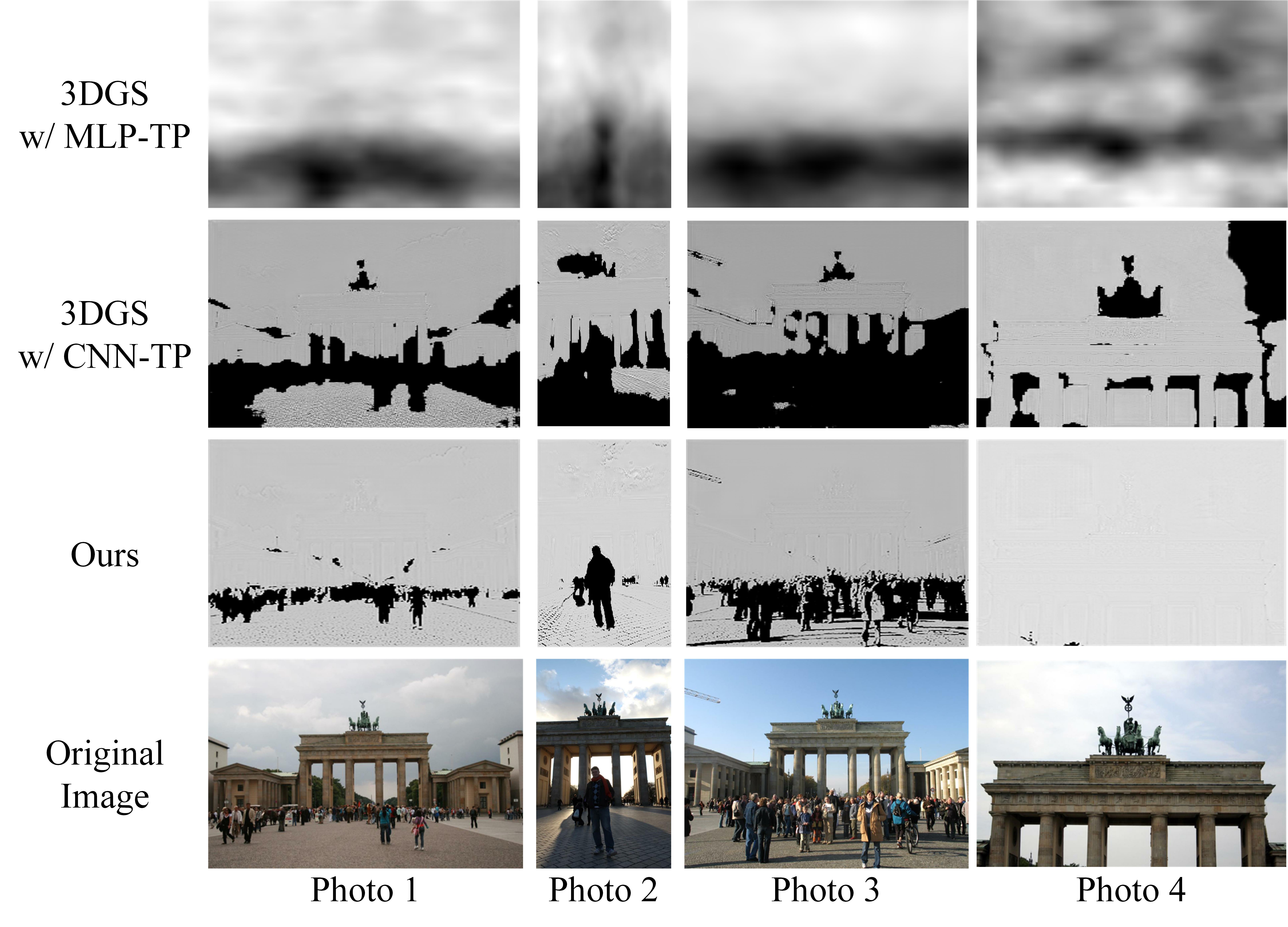}
\caption{Comparisons of masking transient contents between MLP based transient mask predictor (3DGS w/ MLP-TP), CNN based transient mask predictor (3DGS w/ CNN-TP), and the proposed method. }
\label{fig_ablation_mask}
\end{figure}

We validate the design choices of WE-GS on the scene Brandenburg Gate, Sacre Coeur, and Trevi Fountain in PhotoTourism dataset, as shown in Tab. \ref{ablation_study}.

\subsubsection{Influence of the Dimensions of the Appearance Embedding and the Degree of the SH}
The experimental results in rows 1, 2, 3, and 4 of Tab. \ref{ablation_study} and Fig. \ref{ablation_dim} demonstrate that smaller or larger dimensions of the appearance embedding and degree of spherical harmonic undermine the rendering performance. 

\subsubsection{Our model with directly predicting a per-image SH}
In row 5 of Tab. \ref{ablation_study}, we directly predict the SH coefficients for each 3D Gaussian of each unconstrained image, instead of predicting a residual SH. We observed that such a strategy makes it more difficult for the neural network to converge, resulting in a decrease in rendering performance.

\subsubsection{The Influence of Simultaneous Appearance Encoding and Transient Mask Prediction}
In row 6 of Tab. \ref{ablation_study}, we removed the Channel Attention Module (CAM), and we found that both CAM is indispensable for improving the rendering quality. We also assessed the performance of 3DGS with MLP based or CNN based Transient mask Predictor (TP) (mentioned in Sec. \ref{method_encoder}) in rows 7 and 8 of Tab. \ref{ablation_study}. The MLP based mask predictor is implemented similar to the mask predictor in Ha-NeRF \cite{ha-nerf}, and the CNN based mask predictor is implemented similar to our proposed U-Net, except that this U-Net does not act as an attention network but simply predicts the mask of each unconstrained image. Fig. \ref{fig_ablation_mask} demonstrates some visual results. We implemented both the MLP based mask predictor and the CNN based mask predictor with vanilla 3DGS. Both the MLP based mask predictor and the CNN based mask predictor led to poor rendering quality.

\subsubsection{Ablations on the Loss Function Design}

Rows 9 and 10 of Tab. \ref{ablation_study} investigate the effect of the proposed loss function. Without the residual SH regularization loss term $\mathcal{L}_{regSH}$, we found that the base SH of each 3D Gaussian tends to be offset and loses its physical meaning, making it more difficult for converge. With the transient 3D Gaussian regularization loss term $\mathcal{L}_{regTS}$, redundant and transient 3D Gaussians can be effectively pruned, speeding up the optimization process while reducing storage overhead. It is worth noting that the lightweight neural network parameters WE-GS introduced are not the bottleneck, the storage overhead mainly comes from 3D Gaussians.
\section{Conclusion}
In this paper, we introduced WE-GS, an efficient point-based differentiable rendering framework for reconstructing scenes from unconstrained photo collections, accounting for appearance variations and transient objects. Our approach leverages a residual-based SH transfer module to adapt 3DGS to variable lighting and photometric post-processing, enhancing rendering performance without compromise the rendering speed. Additionally, we discovered a symbiotic relationship between the appearance encoder and the transient mask predictor, leading to the design of a lightweight spatial attention module. This module enables simultaneous prediction of transient scene masks and latent appearance representations for each unconstrained image. Our experimental results demonstrate competitive novel view or appearance synthesis quality and low storage costs, alongside real-time rendering efficiency. After precomputing, WE-GS maintains the standard format and rendering pipeline of vanilla 3DGS, allowing for seamless intergration into a broader range of 3DGS applications.

\section*{Acknowledgments}
This paper is supported by National Natural Science Foundation of China (No. 62072020); the Leading Talents in Innovation and Entrepreneurship of Qingdao, China (19-3-2-21-zhc); and Open Project Program of State Key Laboratory of Virtual Reality Technology and Systems, Beihang University (No.VRLAB2024A**).

\bibliographystyle{IEEEtran}
\bibliography{ref}  

\section{Biography Section}
\begin{IEEEbiography}[{\includegraphics[width=1in,height=1.25in,clip,keepaspectratio]{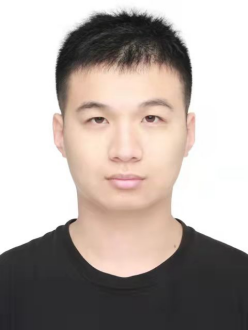}}]{Yuze Wang} is a Ph.D. candidate at State Key Laboratory of Virtual Reality Technology and Systems, School of Computer Science and Engineering, Beihang University, working under the supervision of Prof. Yue Qi. His research cover neural rendering, 3D reconstruction, 3D content generation and editing.
\end{IEEEbiography}
\begin{IEEEbiography}
[{\includegraphics[width=1in,height=1.25in,clip,keepaspectratio]{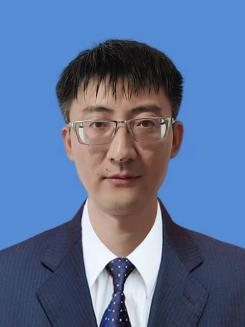}}]{Junyi Wang} obtained his Ph.D. at State Key Laboratory of Virtual Reality Technology and Systems, Beihang University, and now is an assistant processor at School of Computer Science and Technology, Shandong University. His research covers camera localization, object pose estimation and 3D reconstruction.
\end{IEEEbiography}
\begin{IEEEbiography}
[{\includegraphics[width=1in,height=1.25in,clip,keepaspectratio]{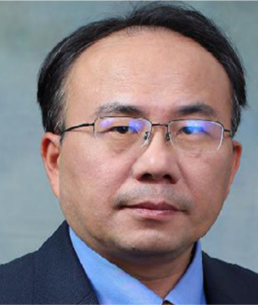}}]{Yue Qi} is a professor at State Key Laboratory of Virtual Reality Technology and Systems, School of Computer Science and Engineering, Beihang University. He was a senior visiting scholar at Harvard University. His research covers Virtual Reality, Augmented Reality algorithms, computer graphics theory and methods, Computer vision algorithms and applications.
\end{IEEEbiography}
\end{document}